\documentclass[a4paper, 10pt, conference]{ieeeconf}      

\IEEEoverridecommandlockouts                              

\usepackage{graphics,color,hyperref} 
\usepackage{epsfig} 
\usepackage{mathptmx} 
\usepackage{times} 
\usepackage{amsmath} 
\usepackage{amssymb}  
\usepackage{cite}
\usepackage{siunitx}
\usepackage{enumerate}
\usepackage[export]{adjustbox}
\usepackage{algorithm}
\usepackage{algorithmicx}
\usepackage{varwidth}
\usepackage[noend]{algpseudocode}
\usepackage{siunitx}
\DeclareMathOperator{\argmin}{arg\,min} 
\DeclareMathOperator{\argmax}{arg\,max} 
\newcommand{\RomanNumeralCaps}[1]
    {\MakeUppercase{\romannumeral #1}}
    
\title{\LARGE \bf
Enhancement of Energy-Based Swing-Up Controller via Entropy Search
}

\author{Chang Sik Lee$^{1}$ and Dong Eui Chang$^{2,3}$
\thanks{$^{1}$School of Electrical Engineering, KAIST, Daejeon, Korea. drancon@kaist.ac.kr}
\thanks{$^{2}$Corresponding author, School of Electrical Engineering, KAIST, Daejeon, Korea. dechang@kaist.ac.kr}
\thanks{$^{3}$This research has been in part supported by KAIST under grant N11180231 and N11190038, and by the ICT R$\And$D program of MSIP/IITP [2016-0-00563, Research on Adaptive Machine Learning Technology Development for Intelligent Autonomous Digital Companion].}}

\begin{document}

\maketitle
\thispagestyle{empty}
\pagestyle{empty}

\begin{abstract}

An energy based approach for stabilizing a mechanical system has offered a simple yet powerful control scheme. However, since it does not impose such strong constraints on parameter space of the controller, finding appropriate parameter values for an optimal controller is known to be hard. This paper intends to generate an optimal energy-based controller for swinging up a rotary inverted pendulum, also known as the Furuta pendulum, by applying  the Bayesian optimization called Entropy Search. Simulations and experiments show that the optimal controller has an improved performance compared to a nominal controller for various initial conditions.

\end{abstract}

\section{INTRODUCTION}

The task of stabilizing an underactuated mechanical system has been investigated over decades. Accordingly, several ideas have been proposed to resolve the problem in improved methods \cite{article:control_lag2,article:StabContLag,article:method_cont,article:energyshaping,article:furuta,article:fradkov}. The idea of using a particular storage function established on the Euler-Lagrange equations of a mechanical system has presented a framework for an effective energy-based swing-up controller \cite{article:shiriaev}. A drawback of the result is that, when it comes to applying it to a real system, the controller requires vague adjustment over a multidimensional parameter space.

Meanwhile, the construction of optimally adjusted controllers has been studied from a wide and diversified point of view \cite{inproceedings:kamwa,article:pso}. In recent years, as the notion of machine learning has been widening its coverage over a variety of fields, it has also begun to put its influence on the  optimal control of mechanical systems\cite{article:autoLQR,article:sc_gp,article:sup_learning_bipedal,article:Reinforcement_learning,article:sample_efficient,article:goal_driven,article:RL_3D_sim}. Da et al.\cite{article:sup_learning_bipedal} deploys supervised learning methods to obtain more robust controllers for a 3D bipedal robot. In \cite{article:Reinforcement_learning} and \cite{article:RL_3D_sim}, reinforcement learning algorithms are used to compensate for unmodeled dynamics of systems. Furthermore, as a sample-efficient methodology to solve non-convex optimization problems, Bayesian optimization are widely adopted to optimize controllers\cite{article:sc_gp,article:sample_efficient,article:goal_driven}.
	
However, all the approaches in \cite{article:autoLQR,article:sc_gp,article:sup_learning_bipedal,article:Reinforcement_learning,article:sample_efficient,article:goal_driven,article:RL_3D_sim} have a common problem that they look for local minima. On the other hand, Marco et al.\cite{article:autoLQR} tackles the task of finding proper parameter values for a controller that optimally stabilizes a linear model by using Entropy Search\cite{article:entropy_search}, a machine learning process which finds a global minimum of a given cost function.

This paper aims to take advantage of the machine learning optimization technique to resolve the drawback of the energy-based control\cite{article:shiriaev} for stabilizing a nonlinear model. To be specific, we use an energy-based controller for a rotary inverted pendulum system, and we intend to fit a Gaussian process estimation model through repeated evaluations of a cost function whose distribution is unknown, following procedures of Entropy Search \cite{article:entropy_search}. Consequently, we can globally estimate the optimal parameter value for the best performance of the controller.

\section{PROBLEM STATEMENT}

Kolesnichenko and Shiriaev  \cite{article:shiriaev}  has proposed an energy-based swing-up controller  for an underactuated mechanical system, and provided sufficient conditions on the controller's gain parameters $K \in \mathbb{R}^{\ell}$ for successful swing-up. However, not all the parameter values under the conditions result in assured swing-up of the real system. Moreover, even though most parameter values can build controllers that drive the system to eventually reach the desired swing-up equilibrium point, their performances may not be all satisfactory. Therefore, there still remains the laborious task to find a set of parameter values which achieves the desired performance  to swiftly reach the desired equilibrium point with less oscillation.

The task to find such values of control parameters is formulated as an optimization problem with a cost function $J(K)$ that properly reflects the desired performance,
\begin{equation}\label{eq:prob_stat_optimal_cost}
    K^{*} = \argmin_{K \in \mathcal{D}} J(K),
\end{equation}
where $\mathcal D$ is a parameter domain.
To solve this optimization problem, we employ  the Bayesian optimization technique called Entropy Search; refer to \cite{article:entropy_search} for more details on Entropy Search. Entropy Search has the merit  that, where not all the values of $J(K)$ are not known, it globally estimates the given cost function $J(K)$ and finds a reliable global minimum while most of  other algorithms seek local minima.

\section{Preliminaries}

Before description of the main result, we offer backgrounds on Entropy Search.

\subsection{Entropy Search} 

The problem \eqref{eq:prob_stat_optimal_cost} can be stated as finding $K^{*} \in \mathcal{D}$ that optimizes a function $J(K)$ while the functional relationship between $K$ and $J(K)$ is not known a priori. Namely, the values of cost function $J(K)$ may not be available or observable for all  $K \in \mathcal D$.  In such a situation, Bayesian optimization methods are quite useful since they repeatedly estimate an arbitrary black box function ``$J(K)$" based on a probabilistic model and selects an appropriate measure point $K_{\rm next}$ for more accurate modeling.  Among several available Bayesian techniques, we choose to use Entropy Search which efficiently finds global minimum  \cite{article:entropy_search}.


Two tools are required for  Bayesian optimization. One is a probabilistic model for estimating the black box function $J(K)$ based on measurements 
\begin{equation}\label{eq:Hn}
H_{n}= \left\{J(K_{1}), J(K_{2}), \ldots, J(K_{n})\right\},
\end{equation}
 and the other is a decision rule for specifying a new point $K_{n+1}$ where $J(K_{n+1})$ will be evaluated so that the estimation model approaches closer to the actual values of $J(K)$.

First, as its estimation model, Entropy Search utilizes a Gaussian process. A Gaussian process is a non-parametric model generally used to estimate an unknown function $J(K)$. Suppose $m(K)$ as a prior mean and $k(K_{j}, K_{l})$ as a covariance function (kernel) between $J(K_{j})$ and $J(K_{l})$, where $K_{j}, K_{l} \in \mathcal{D}$. The former implies the prior belief on $J(K)$, which is usually a constant, and the latter suggests the relationship between those two random variables $J(K_{j})$ and $J(K_{l})$.
Given a set of evaluation \eqref{eq:Hn} at a set of points given by
\begin{equation}\label{eq:hn}
h_{n} = \left\{K_{1}, K_{2}, \ldots, K_{n} \right\},
\end{equation}
the function value $J(K_{\rm new})$ at a new point $K_{\rm new}$ is a random variable with a Gaussian distribution with the posterior mean and variance given respectively by
\begin{align*}
&\mu_{n}(K_{\rm new}) = m(K_{\rm new}) + \mathbf{k}_{n}(K_{\rm new})\mathbf{K}_{n}^{-1}y_{n},\\
&\sigma_{n}^{2}(K_{\rm new}) = k(K_{\rm new},K_{\rm new}) - \mathbf{k}_{n}(K_{\rm new})\mathbf{K}_{n}^{-1}\mathbf{k}^{T}_{n}(K_{\rm new}),
\end{align*}
where 
\begin{align*}
&\left[\mathbf{K}_{n}\right]_{ij} = k(K_{i},K_{j}) \quad i,j \in \left\{1, 2, 3, \ldots,  n \right\},\\
&\mathbf{k}_{n}(K_{\rm new}) = \left[k(K_{\rm new},K_{1}), k(K_{\rm new},K_{2}),\ldots,  k(K_{\rm new},K_{n})\right],\\
&y_{n} = \left[(J(K_{1}) - m(K_{1}), J(K_{2}) - m(K_{2}), \ldots, J(K_{n}) - m(K_{n}) \right]^{T}.
\end{align*}
Utilization of above equations allows us to estimate the functional relationship between $J(K)$ and $K$. For more details, refer to \cite{article:entropy_search,article:MF_ES}

Secondly, in order to determine the next measurement point, Entropy Search computes  the expected change $E[\bigtriangleup \mathbf{H}]$ in entropy $\mathbf{H}$ of $P_{\rm min}$, where $P_{\rm min}$ and $\mathbf{H}$  are defined as
\begin{align*}
 P_{\rm min}(K) &= P(K = \argmin_{K' \in \mathcal{D}}\hat{J}(K')),\\
 \mathbf{H}(K) &= \int_{\mathcal{D}} P_{\rm min}(K)\log(\frac{P_{\rm min}(K)}{U(K)}) dK,
\end{align*}
with $\hat{J}(K)$ being the Gaussian process estimation of $J(K)$. i.e. $\hat{J}(K) \sim \mathcal{N}(\mu_{n}(K),\sigma^{2}_{n}(K)) \enspace \forall K \in \mathcal{D}$, and $U(K)$ is the uniform distribution over $\mathcal{D}$.
The next measurement point $K_{n+1}$ is then selected by finding a point with the largest expected change in entropy ($E[\bigtriangleup \mathbf{H}]$). This decision rule  is established on the assumption that the next measurement point $K_{n+1}$ obtained as above is the most informative point. 

The measurement of $J(K_{n+1})$ is  made at the new point $K_{n+1}$, and then $J(K_{n+1})$ and $K_{n+1}$ are added respectively to the sets $H_{n}$ and $h_n$ after which the two sets are renamed as $H_{n+1}$ and $h_{n+1}$. 
 Entropy Search then returns a best guess point $K_{\rm bg}$ at which the cost function $J(K)$ is likely to be minimum, that is, where $P_{\rm min}$ is the largest by definition of $P_{\rm min}$. This makes the end of a single process. 

The process is repeated until the model has sufficiently converged to the objective function $J(K)$ and $P_{\rm min}$ is peaked around the optimum \cite{article:MF_ES}. Namely, the termination of the process is determined when a posterior mean at a best guess $\mu_{n}(K_{\rm bg})$ does not change over a threshold $\epsilon$ for  $\gamma$ consecutive  iterations. For more details including derivation of $E[\bigtriangleup \mathbf{H}]$, refer to \cite{article:entropy_search}.

To sum up, given an initial condition, a termination threshold $\epsilon$, a duration $\gamma$, and a set of evaluations \eqref{eq:Hn} at arbitrary points \eqref{eq:hn}, Entropy Search can be described as in the following algorithm:
\begin{algorithm}
\caption{Entropy Search \cite{article:autoLQR}}\label{alg:ES}
\begin{algorithmic}[1]
\Procedure{Entropy Search}{$m,k,H_{n},h_{n}$}
\State \Comment{$m$ : prior mean, $k$ : kernel function}
\State \Comment{$H_{n}$ in \eqref{eq:Hn}, $h_{n}$ in \eqref{eq:hn}}
\For{$i = 1 \text{ to } N$}

\State \begin{varwidth}[t]{\linewidth}
	$\text{Compute}\enspace (\mu_{i}(K), \sigma^{2}_{i}(K))$ $\forall K\in \mathcal{D}$\par
\hskip\algorithmicindent $\text{with} \enspace GP(m,k,H_{n+i-1},h_{n+i-1}))$\par
    
\hskip\algorithmicindent  \Comment{$\mu_{i}(K)$ : GP posterior mean }\par
\hskip\algorithmicindent  \Comment{$\sigma^{2}_{i}(K)$ : GP posterior variance}
\end{varwidth}
\State \begin{varwidth}[t]{\linewidth}
$P_{{\rm min},i}(K) \gets \text{Compute} \enspace P_{\rm min}(\mu_{i}(K), \sigma^{2}_{i}(K))$\par
\hskip\algorithmicindent \qquad \qquad \qquad \qquad \qquad \qquad \qquad \quad $\forall K\in \mathcal{D}$
\end{varwidth}
\State $\mathbf{H}_{i}(K) \gets \text{Compute}\enspace \mathbf{H}(P_{{\rm min},i}(K))\enspace \forall K\in \mathcal{D}$
\State $E[\bigtriangleup \mathbf{H}_{i}(K)] \gets \text{Compute} \enspace E[\bigtriangleup \mathbf{H}_{i}(K)]\enspace \forall K\in \mathcal{D}$
\State $K_{n+i} \gets \argmax_{K \in \mathcal{D}} E[\bigtriangleup \mathbf{H}_{i}(K)]\enspace \forall K\in \mathcal{D}$
\State $\text{Generate a controller } u(K_{n+i})$
\State $\text{Run a simulation or an experiment with } u(K_{n+i})$
\State $\text{Compute } J(K_{n+i})$
\State $(H_{n+i}, h_{n+i}) \gets (H_{n+i-1}, h_{n+i-1}) \cup (J(K_{n+i}), K_{n+i})$
\State $K_{\rm bg} \gets \argmax_{K \in \mathcal{D}}P_{\rm min}(K)$
\If{$\| \mu_{i}(K_{\rm bg}) - \mu_{i-j}(K_{\rm bg})| < \epsilon$\\
	\hskip\algorithmicindent\hskip\algorithmicindent\hskip\algorithmicindent $\forall j \in \left\{1,2,\ldots,\gamma-1 \right\}$}
\State $\text{break}$
\EndIf
\EndFor
\State 
\Return $K_{\rm bg}$
\EndProcedure
\end{algorithmic}
\end{algorithm}

\begin{figure}[t]
	\includegraphics[trim={0cm 0cm, 0cm, 0cm},clip,width = 0.45\textwidth,height = 0.285\textheight]{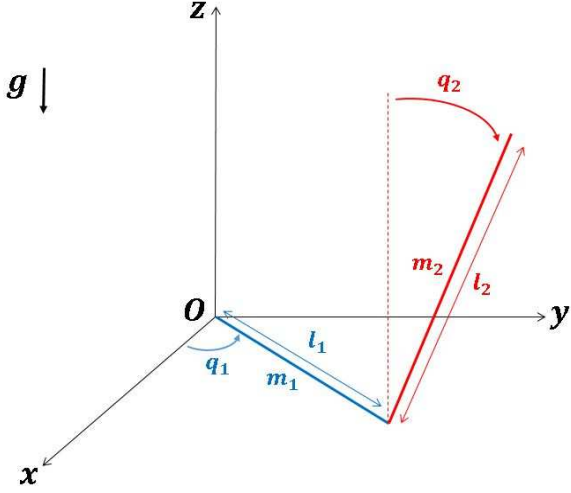}
	\caption{A simplified drawing of QUBE Servo2.}
	\label{fig:system_drawing}
\end{figure}

\section{Swing Up of the Furuta Pendulum}
\subsection{Swing-Up Controller}

As an underactuated mechanical system, we choose  Quanser QUBE Servo 2\cite{manual:quanser_manual} which is a kind of  Furuta pendulum. Assume an ideal model of the Furuta pendulum system with no noise and no frictions. The  configuration space $Q$ of the system is $Q= \mathbb{R} \times \mathbb{R}$, $q = (q_{1}, q_{2})$ where $q_{1}$ is an angle of the rotary arm, $q_{2}$ is an angle of the inverted pendulum, as shown in Figure \ref{fig:system_drawing}.
The Lagrangian $\mathcal{L}$ of the  system is given by
\begin{align*}
\mathcal{L}(q,\dot{q}) = \frac{1}{2}\dot{q}^{T}M(q)\dot{q} - PE(q)
\end{align*}
where 
\begin{align*}
&M(q) = \left(\begin{array}{cc} I_{10} + I_{11}\sin^{2}{q_{2}} & -I_{12}\cos{q_{2}} \\ -I_{12}\cos{q_{2}} & I_{2} \end{array}\right),\\
&PE(q) = V_{0}\cos{q_{2}}
\end{align*}
with $PE(q)$ being the potential energy.
The Euler-Lagrange equations of the system are computed as
\begin{align*}
    M(q)\left(\begin{array}{c} \ddot{q}_{1} \\ \ddot{q}_{2} \end{array}\right) + C(q,\dot{q})\left(\begin{array}{c} \dot{q}_{1} \\ \dot{q}_{2} \end{array}\right) + 
    G(q,\dot{q}) = \left(\begin{array}{c} u \\ 0 \end{array}\right)
\end{align*}
where
\begin{align*}
C(q,\dot{q}) &= \left(\begin{array}{cc} 2I_{11}\dot{q}_{2}\sin{q_{2}}\cos{q_{2}} & I_{12}\dot{q}_{2}\sin{q_{2}} \\ -I_{11}\dot{q}_{1}\sin{q_{2}\cos{q_{2}}} & 0\end{array}\right), \\
G(q,\dot{q}) &= \left(\begin{array}{c} 0 \\ -V_{0}\sin{q_{2}} \end{array}\right),
\end{align*}
and
\begin{align*}
I_{10} &= J_{1} + m_{2}l_{1}^{2},\enspace I_{11} = \frac{m_{2}l_{2}^{2}}{3},\enspace  I_{12} = \frac{m_{2}l_{1}l_{2}}{2},\\
I_{2} &= J_{2} + \frac{m_{2}l_{2}^{2}}{4}, \enspace V_{0} = \frac{m_{2}l_{2}g}{2},
\end{align*}
where $m_{1}$ and $m_{2}$ are masses, $J_{1}$ and $J_{2}$ are moments of inertia, $l_{1}$ and $l_{2}$ are lengths of rotary arm and pendulum respectively. The symbol $g$ denotes the gravitational acceleration and $V_{0}$ is the potential energy at the equilibrium point $(q_1, q_2, \dot q_1, \dot q_2) = (0,0,0,0)$.  The values of the parameters are
\begin{align*}
m_1 &= 0.095\enspace \si{kg},\enspace m_2 = 0.024\enspace \si{kg},\enspace g = 9.81\enspace \si{m/s^{2}},\\
l_1 &= 0.085\enspace \si{m},\enspace J_1 = 5.72 \times 10^{-5}\enspace \si{kg\cdot m^2},\\
l_2 &= 0.129\enspace \si{m},\enspace  J_2 = 3.33 \times 10^{-5}\enspace \si{kg\cdot m^{2}},
\end{align*}
which are from  the table on p.8 of \cite{manual:quanser_manual}. The total energy $E$ is given by 
\begin{align*}
E(q,\dot{q}) &= \frac{1}{2}\dot{q}^{T}M(q)\dot{q} +  PE(q)  \\
&=\frac{1}{2}((I_{10} + I_{11}\sin^{2}{q_{2}})\dot{q}_{1}^{2} + I_{2}\dot{q}_{2}^{2}) 
- I_{12}\dot{q}_{1}\dot{q}_{2}\cos{q_{2}} \\
&\quad+ V_{0}\cos{q_{2}}.
\end{align*}
Kolesnichenko and Shiriaev  \cite{article:shiriaev}  introduces the following storage function $V(q,\dot q)$:
\begin{equation}\label{eq:storage_exp}
V(q,\dot{q}) = k_{E}\frac{1}{2}(E - E_{0})^{2} + k_{v}\frac{1}{2}\dot{q}_{1}^{2} + k_{x}(1-\cos{q_{1}}),
\end{equation}
where the original term   $q_{1}^2/2$ in Kolesnichenko and Shiriaev  \cite{article:shiriaev} has been replaced by $(1-\cos q_1)$  in order to   take  the periodicity of angle into account. From the storage function, one can easily derive the following energy-based controller 
\begin{equation}\label{eq:u_derivation}
    u = \frac{-k_{p}\dot{q}_{1}-H(q,\dot{q})}{k_{E}(E - E_{0})+k_{v}R(q)}
\end{equation}
where 
\begin{align*}
     H(q,\dot{q}) &= -k_{v}\left(\begin{array}{cc} 1 & 0 \end{array}\right)M(q)^{-1} C(q,\dot{q})\left(\begin{array}{c} \dot{q}_{1} \\ \dot{q}_{2} \end{array}\right)\\ 
     &\quad-k_{v}\left(\begin{array}{cc} 1 & 0 \end{array}\right)M(q)^{-1}G(q,\dot{q}) + k_{x}(1-\cos{q_{1}}),\\
     R(q)&=\frac{I_{2}}{(I_{10}I_{2} - I_{12}^{2}\cos^2q_{2} + I_{11}I_{2}\sin^{2}{q_{2}})}.
\end{align*}
See  \cite{article:shiriaev} for a detailed derivation. 

The swing-up control law \eqref{eq:u_derivation} contains 4 parameters: $k_{p}$, $k_{E}$, $k_{v}$ and $k_{x}$, which are put in vector form as follows:
\[
K =(k_{p}, k_{E}, k_{v}, k_{x}) \in \mathbb R^4.
\]
According to Theorem 2 of \cite{article:shiriaev}, a sufficient condition on $K$ for successful swing-up is given by
\begin{equation}\label{eq:constraint}
k_{v} > 6.8366\times10^{-6}k_{E}.
\end{equation}
In the range of $|q_{2}| \leq \ang{20}$, the swing-up controller  \eqref{eq:u_derivation} is switched to the LQR for the linearization of the system at the equilibrium point with the weight matrices   $Q = \operatorname{diag}([1, 10, 1, 10])$ and $ R = 10000$.

To sum up, we swing up  the rotary inverted pendulum relying on the energy-based controller \eqref{eq:u_derivation}. When the pendulum is in the region where the linearized model is effective, the LQR is turned on to hold the pendulum at the desired equilibrium point.

\subsection{Optimization of Swing-Up Controller via Entropy Search}

This section explains practical details about the optimization task to obtain an optimal swing-up controller. We first provide a common setup for simulations and experiments such as the range of parameters, a cost function, and the initial condition. The range of parameter vector $K$ is set as
\begin{align}\label{eq:param_range}
    &400 \leq k_{p} \leq 900,\enspace 10^{6} \leq k_{E} \leq 10^{7},\\
    &5 \leq k_{v} \leq 100,\enspace 100 \leq k_{x} \leq 1000,
\end{align} 
which defines the bounded domain $\mathcal{D}$.
The above range  for $K$  is determined on the basis of the following observations: In the controller formula \eqref{eq:u_derivation}, the energy term $(E-E_0)$ is relatively small due to the small values of the system's physical parameters, so  the gain $k_{E}$ to the energy term is chosen from  the range, $10^{6} \leq k_{E} \leq 10^{7}$, of large numbers relative to other gains. Moreover, the controller has a tendency to work well when $k_{v}$ is close to its lower bound $6.8366\times10^{-6}k_{E}$ given in  \eqref{eq:constraint}, from which the range, $5 \leq k_{v} \leq 100$, is derived.  Ranges of the other parameters $k_{p}$ and $k_{x}$ are chosen in a way that the controller works well, provided that $k_{E}$ and $k_{v}$ are readily set in the above ranges, in several simulations.

We set a cost function as follows:
\begin{align}\label{eq:cost_J2}
&J(K) = \int_{t_{0}}^{t_{f}} \left[\frac{20(1 - \cos{x_{1}(t)})}{5 - \cos{x_{1}(t_{0})}}+\frac{100(1 - \cos{x_{2}(t)})}{30 - \cos{x_{2}(t_{0})}}\right. \nonumber \\ 
&\left. + \frac{1}{2}\left (\frac{\dot{x}_{1}(t)}{80 + |\dot{x}_{1}(t_{0})|} \right )^{2}
+\frac{1}{2}\left(\frac{\dot{x}_{2}(t)}{100 + |\dot{x}_{2}(t_{0})|} \right)^{2}\right] dt,
\end{align}
where $t_0$ is the initial time, $t_f$ is the terminal time, and we use the following state vector
\[
x = (x_1, x_2, x_3, x_4) = (q_1, q_2, \dot q_1, \dot q_2).
\]
By introducing initial conditions in denominators, the cost value defined in \eqref{eq:cost_J2}   is less influenced by modification of initial conditions, which makes cost values comparable over various initial conditions. For these reasons, \eqref{eq:cost_J2} is used to measure performance of the controller in this paper.

The default initial condition for  simulations and  experiments is set as
\begin{equation}\label{default:IC}
x_{0} = \left  (0, \enspace \frac{7\pi}{9}, \enspace 0,\enspace 0 \right ).
\end{equation}

With the setting given above, we find a nominal controller $u(K_{\rm nom})$ by running 10,000 simulations in Matlab Simulink, where the time span of each simulation is 30 seconds.  Each  simulation starts with choosing a gain parameter vector $K = (k_{p}, k_{E}, k_{v}, k_{x})$ uniformly randomly from the range \eqref{eq:param_range}, and ends with computing a cost value $J(K)$. After all the simulations are finished, the set of parameter vectors which result in the lowest costs in the simulations are tested in experiments  to obtain their experimental costs. Through this procedure, a set of parameter values which yields the lowest experimental cost has been found as follows:
\[
K_{\rm nom} = (770.152,\enspace 6255313.438,\enspace 35.190,\enspace 465.098),
\]
which is used as the nominal parameter vector.  

We now find an optimal controller $u(K_{\rm bg})$ using Entropy Search. For Gaussian process, we choose constant prior mean $m(K) = 20$ and the rational quadratic kernel function
\[
k(K_j,K_l) = s^2 \left ( 1 + \frac{1}{2\alpha} (K_j - K_l)S^{-1}(K_j-K_l)\right )^{-\alpha}
\]
with $s^{2} = 9.894$,  $\alpha = 0.131$ and 
\[
S = \operatorname{diag}(\left[58.552,\enspace 40.343,\enspace 21.515,\enspace 271.180\right]).
\]
The hyperparameters, $m(K)$, $s$, $\alpha$ and $S$, for the Gaussian process have been determined based on the result of running several times of simulations and hyperparameter fittings\cite{article:MF_ES}.
\begin{figure}[t]
	\centering
	\includegraphics[trim={0cm 0cm 0cm, 0cm},clip,width = 0.45\textwidth]{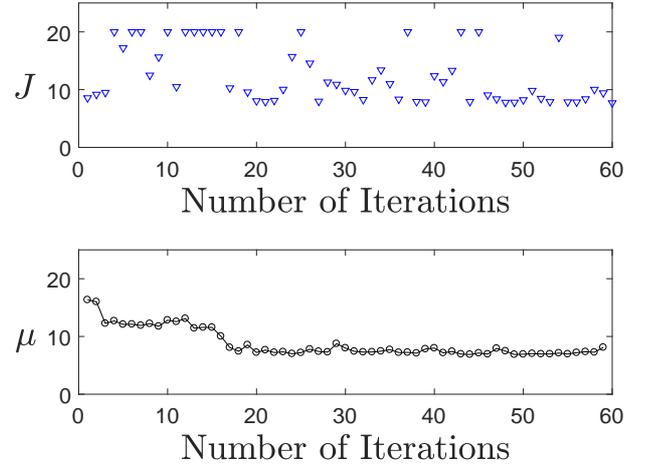}
	\caption{The upper plot shows the cost value $J(K_{n+i})$ evaluated in  line 12 of Algorithm \ref{alg:ES} for each iteration of Entropy Search. The lower plot shows a posterior mean ($\mu_{n}(K_{\rm bg})$, black) at a best guess $K_{\rm bg}$ computed in  line 14 of Algorithm \ref{alg:ES} for each iteration of Entropy Search.}
	\label{fig:ES_process}
\end{figure}
Before initializing Algorithm \ref{alg:ES} to perform Entropy Search, we run 5 simulations with the default initial condition \eqref{default:IC} to form a set of initial observations $H_{5}$ \eqref{eq:Hn} at a set of points $h_{5}$ \eqref{eq:hn}. Once the sets $H_{5}$ and $h_{5}$ are made, Entropy Search starts by running Algorithm \ref{alg:ES}.
We use simulations, in line 11 of Algorithm \ref{alg:ES}, to compute trajectories of the system driven by  controller $u(K_{n+i})$ where a single simulation is run for 30 seconds with the default initial condition. The process is terminated when the posterior mean $\mu_{n}(K_{\rm bg})$ at the best guess $K_{\rm bg}$ has not changed more than $\epsilon= 0.01$ for $\gamma= 3$ iterations or when an iteration is repeated for $N= 60$ times. Verification of the resultant controller $u(K_{\rm bg})$ is executed in a simulation and an experiment for 30 seconds after Algorithm \ref{alg:ES} is completed.
\begin{figure}[b]
	\includegraphics[trim={0cm 0cm 0cm, 0cm},clip,width = 0.47\textwidth,height = 0.27\textheight]{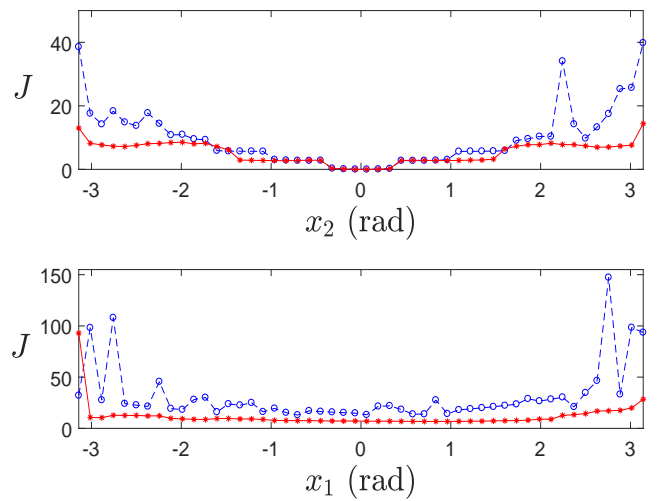}
	\caption{ Simulation cost values of the optimal controller $u(K_{\rm ES})$ (solid red) and the nominal controller $u(K_{\rm nom})$ (dotted blue) for various initial conditions. The upper plot is cost values $J$ evaluated over the initial conditions $x_{1}(0) = 0$, $-\pi \leq x_{2}(0) \leq \pi$, and zero initial velocity. The lower plot is cost values $J$ evaluated over the initial conditions $-\pi \leq x_{1}(0) \leq \pi$, $x_{2}(0) = \frac{5\pi}{6}$, and zero initial velocity. Most cost values of the optimal controller $u(K_{\rm ES})$ are lower than cost values of the nominal controller $u(K_{\rm nom})$.}
	\label{fig:init_var_sim}
\end{figure}

After 60 iterations, Entropy Search obtains the optimal parameter vector 
\[
K_{\rm ES} = (467.727, \enspace 3015436.481 , \enspace 13.235,\enspace 273.014).
\]
Figure \ref{fig:ES_process} shows how Entropy Search has converged to $K_{\rm ES}$ by iteratively evaluating a cost value $J(K_{n+i})$ and estimating a posterior mean $\mu_{n}(K_{\rm bg})$ at a best guesses $K_{\rm bg}$. To be specific, in the upper side of Figure \ref{fig:ES_process}, a cost value $J(K_{n+i})$ obtained at a suggested point $K_{n+i}$, following the line 9 -- 12 of Algorithm \ref{alg:ES}, is plotted for each iteration. In the lower side, a posterior mean $\mu_{n}(K_{\rm bg})$ at a best guess $K_{\rm bg}$ given in  line 14 of Algorithm \ref{alg:ES} is plotted for each iteration. As the iterative process goes on, the posterior mean $\mu_{n}(K_{\rm bg})$ at the best guess point $K_{\rm bg}$ approaches to a certain value, which indicates that the estimation model has been fit to the real distribution of $J(K)$ over iterations.
\begin{figure}[t]
	\includegraphics[trim={0cm 0cm 0cm, 0cm},clip,width = 0.47\textwidth,height = 0.22\textheight]{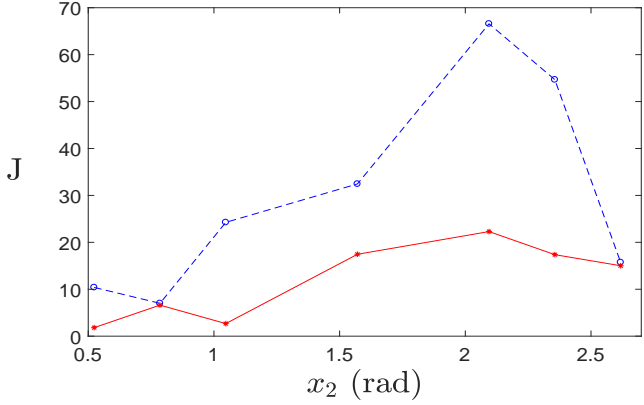}
	\caption{Experimental cost comparison between the optimal controller $u(K_{\rm ES})$ (solid red) and the nominal controller $u(K_{\rm nom})$ (dotted blue) over the initial conditions $x_{2}(0)$ in \eqref{eq:x2_initial}. All the cost values of the optimal controller $u(K_{\rm ES})$ are lower than cost values of the nominal controller $u(K_{\rm nom})$.}
	\label{fig:init_var_real}
\end{figure}
\begin{figure}[b]
	\includegraphics[trim={0cm 0cm 0cm, 0cm},clip,width = 0.47\textwidth,height = 0.28\textheight]{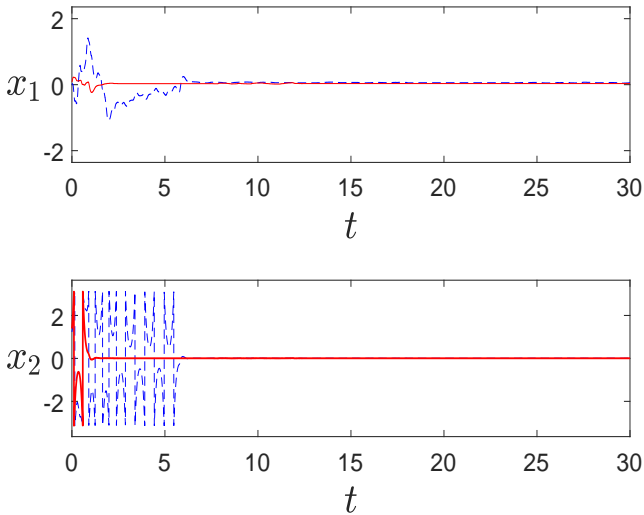}
	\caption{Experimental time response comparison between the optimal controller $u(K_{\rm ES})$ (solid red) and the nominal controller $u(K_{\rm nom})$ (dotted blue) at initial $x_{2}(0)$ = $\frac{\pi}{3}$.}
	\label{fig:pi3}
\end{figure}
\begin{figure}[t]
	\includegraphics[trim={0cm 0cm 0cm, 0cm},clip,width = 0.47\textwidth,height = 0.28\textheight]{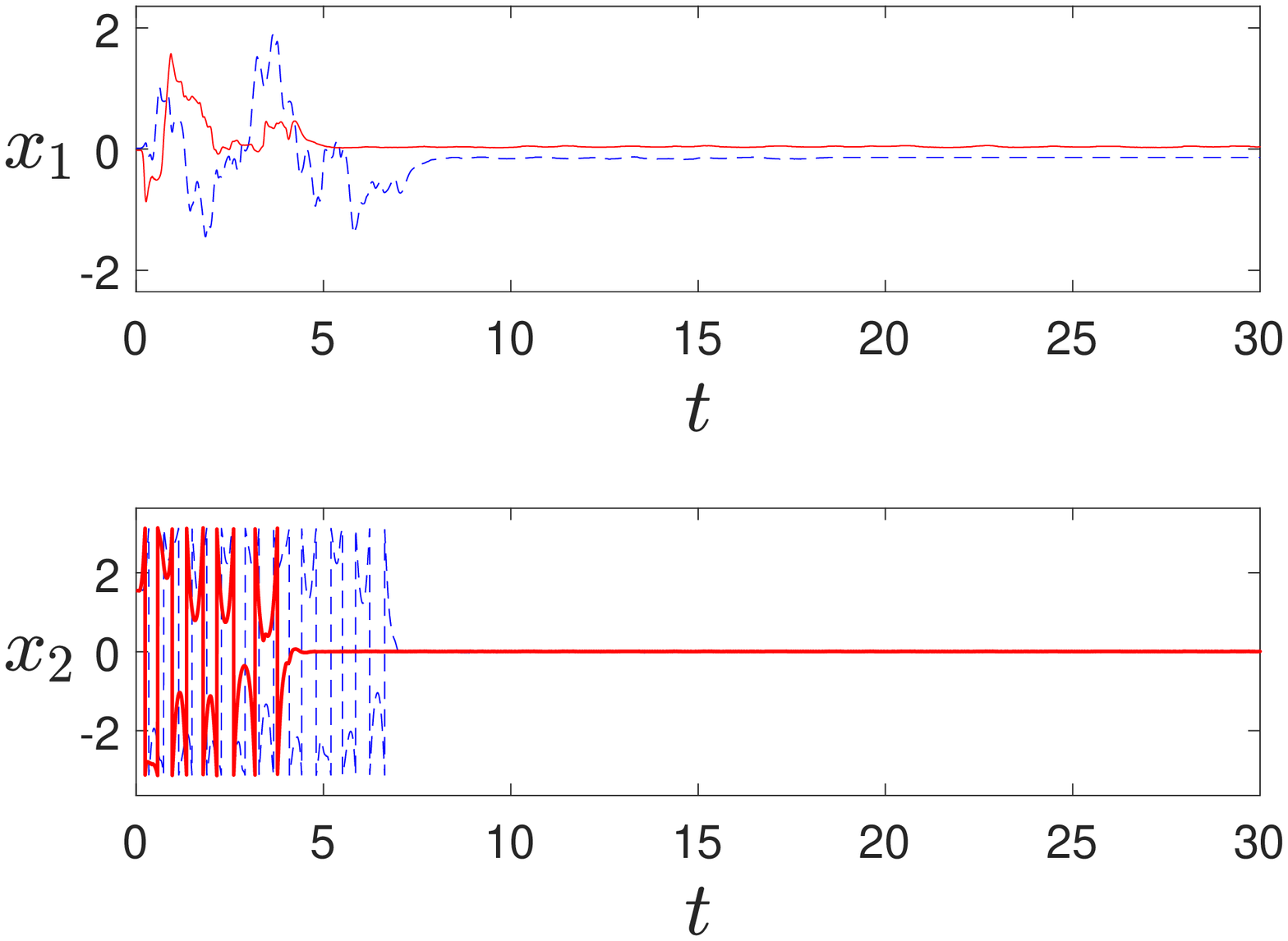}
	\caption{Experimental time response comparison between the optimal controller $u(K_{\rm ES})$ (solid red) and the nominal controller $u(K_{\rm nom})$ (dotted blue) at $x_{2}(0)$ = $\frac{\pi}{2}$ where the other initial states are zero.}
	\label{fig:pi2}
\end{figure}
\begin{figure}[b]
	\includegraphics[trim={0cm 0cm 0cm, 0cm},clip,width = 0.47\textwidth,height = 0.28\textheight]{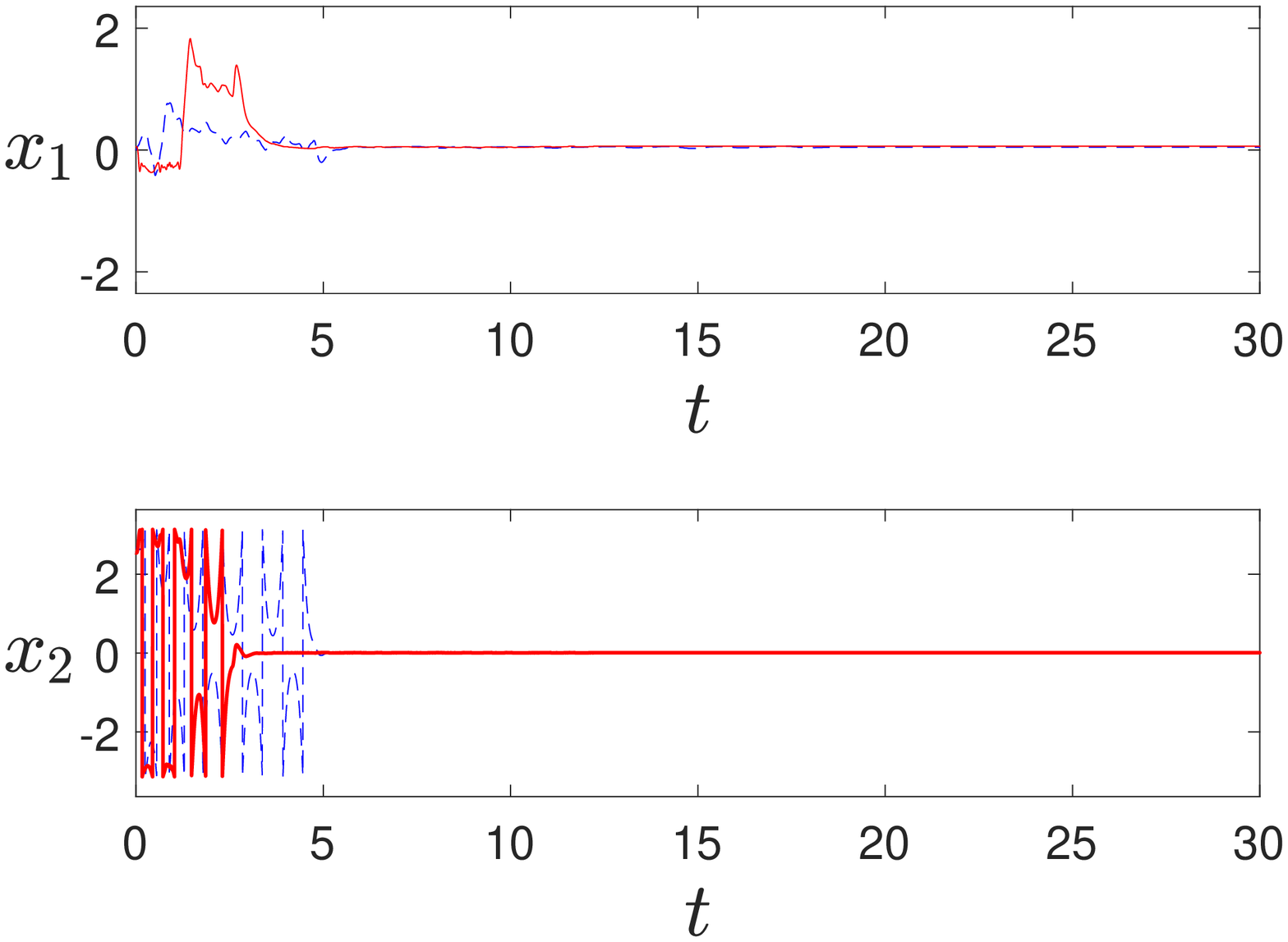}
	\caption{Experimental time response comparison between the optimal controller $u(K_{\rm ES})$ (solid red) and the nominal controller $u(K_{\rm nom})$ (dotted blue) at $x_{2}(0)$ = $\frac{5\pi}{6}$ where the other initial states are zero.}
	\label{fig:5pi6}
\end{figure}

\subsection{Performance Comparison}

We have run two simulations for the default initial condition \eqref{default:IC}: one with the nominal controller $u(K_{\rm nom})$ and the other with the optimal controller $u(K_{\rm ES})$, and have obtained the following cost values:
\[
J(K_{\rm nom}) = 12.286, \quad J(K_{\rm ES}) = 8.954
\] 
from which it is deduced that the optimal controller yields a cost value $27.12 \%$ than the nominal controller.  Although the optimal gain $K_{\rm ES}$ has been obtained for the default initial condition, our exhaustive simulations show that it performs well for various initial conditions in the range of $-\pi \leq x_{1} \leq \pi$, $-\pi \leq x_{2} \leq \pi$ with zero initial velocity. Figure \ref{fig:init_var_sim} shows cost values of the optimal controller $u(K_{\rm ES})$ and the nominal controller $u(K_{\rm nom})$ sampled from the set of entire costs computed in simulations, where they respectively form plots over initial conditions.

For the purpose of verification, we test the two controllers $u(K_{\rm nom})$ and $u(K_{\rm ES})$ on the system of Quanser QUBE Servo 2 for the following  initial conditions:
\begin{align}\label{eq:x2_initial}
x_2(0) \in \left\{\frac{\pi}{6},\enspace \frac{\pi}{4},\enspace \frac{\pi}{3},\enspace \frac{\pi}{2},\enspace \frac{2\pi}{3},\enspace \frac{3\pi}{4},\enspace \frac{5\pi}{6}\right\}
\end{align}
with the other states at zero. 

For each initial condition, the cost value $J$ is computed by averaging the cost values of 5 repeated experiments. The results are plotted in Figure \ref{fig:init_var_real}. It can be seen that the optimal controller produces a lower cost value for each initial condition than the nominal controller.  The time responses of the two controllers for the initial conditions $x_2(0)  \in \left\{\frac{\pi}{3},\enspace \frac{\pi}{2},\enspace \frac{5\pi}{6}\right\}$ are measured in experiments and plotted in Figures \ref{fig:pi3}, \ref{fig:pi2}, and \ref{fig:5pi6}, repectively. It can be seen that the response with the optimal controller $u(K_{\rm ES})$ has a shorter settling time than the nominal controller for each initial condition. It follows that Entropy Search has succeeded in isolating an energy-based controller with the best performance, which leads to quick and firm stabilization of the rotary inverted pendulum. The video of the experiments is available at \url{https://youtu.be/JcmpLU5rJCg}.

\section{CONCLUSIONS}

The energy based controller proposed in \cite{article:shiriaev} is not only derived easily by considering the energy of system but also effective in stabilizing an underactuated non-linear system. However, it still requires a considerable amount of efforts, such as searching through multidimensional hyper-parameter space, to isolate optimal parameter values. This paper proposes application of Entropy Search to the problem of finding the optimal gain parameter values of an energy-based swing-up controller for the Furuta pendulum system. Based on the results in Section \RomanNumeralCaps{4-C}, it is concluded that Entropy Search successfully optimizes the given controller so that the optimal controller attains a better performance than the nominal controller.  In the future, we will combine Entropy Search with a deep neural network \cite{Ch17book} to enhance the performance of the controller. 

\addtolength{\textheight}{-12cm}   


\end{document}